\documentclass[a4paper,11pt]{article}
\usepackage{head,fullpage}     
\usepackage[pdftex]{graphicx}  
\usepackage{url}
\usepackage{datetime}

\usepackage[numbers]{natbib}

\usepackage[table,xcdraw]{xcolor}
\usepackage{float}
\usepackage{algorithm}
\usepackage{algorithmic}
\usepackage{mathtools}
\usepackage{hyperref}
\usepackage{cancel}

\newdateformat{monthyeardate}{%
  \monthname[\THEMONTH] \THEYEAR}

\title{\textbf{A contextual analysis of multi-layer perceptron models  in classifying hand-written digits and letters: limited resources}}

\author{
        \textbf{Tidor-Vlad Pricope} \\
        The University of Edinburgh \\
        Informatics Forum, Edinburgh, UK, EH8 9AB \\
        \texttt{T.V.Pricope@sms.ed.ac.uk} \\
}

\begin{document}
\date{}
\maketitle

\begin{abstract}
        Classifying hand-written digits and letters has taken a big leap with the introduction of ConvNets. However, on very constrained hardware the time necessary to train such models would be high. Our main contribution is twofold. First, we extensively test an end-to-end vanilla neural network (MLP) approach in pure numpy  without any pre-processing or feature extraction done beforehand. Second, we show that basic data mining operations can significantly improve the performance of the models in terms of computational time, without sacrificing much accuracy. We illustrate our claims on a simpler variant of the Extended MNIST dataset, called Balanced EMNIST dataset. Our experiments show that, without any data mining, we get increased generalization performance when using more hidden layers and regularization techniques, the best model achieving 84.83\% accuracy on a test dataset. Using dimensionality reduction done by PCA we were able to increase that figure to 85.08\% with only 10\% of the original feature space, reducing the memory size needed by 64\%. Finally, adding methods to remove possibly harmful training samples like deviation from the mean helped us to still achieve over 84\% test accuracy but with only 32.8\% of the original memory size for the training set. This compares favorably to the majority of literature results obtained through similar architectures. Although this approach gets outshined by state-of-the-art models, it does scale to some (AlexNet, VGGNet) trained on 50\% of the same dataset.
\end{abstract}

\section{Introduction}

The different possible architectures of a neural network play a key role in achieving success with them. There is a plethora of different neural network methods and architectures that are used in the literature, Convolution and Capsule based Neural Networks being on the basis of the current state of the art in \textbf{computer vision} \cite{patrick2019capsule}. Just over a span of a few years, the top-5 image classification accuracy over the ImageNet dataset has increased from \textbf{84\%} \cite{krizhevsky2017imagenet} to \textbf{95\%} \cite{szegedy2015going, simonyan2014very}, using deeper networks with rather small receptive fields \cite{ciregan2012multi}. However, training or using such complex models on less capable machines like a Raspberry Pi with limited memory and computational power is not trivial because of the extreme number of floating-point operations necessary. 

This paper investigates the impact of vanilla shallow and deeper neural networks models with basic data mining and varying hyper-parameters on building a strong digit and letter classifier. We aim to find the best trade-off between the memory, computational time needed and accuracy for a multi-layer perceptron model. We show that deeper multi-layer perceptron models with regularization mechanisms are much better than the other tested in this paper and performance can be even increased with data mining techniques like PCA (Principal Component Analysis) \cite{wold1987principal}. This is to be expected as there seems to be a general rule that deeper is better and other results in this area have also underscored the superiority of deeper networks \cite{yu2013feature} in terms of accuracy and/or performance. In addition, we are also going to test and compare performances of the MLP models with other popular non-neural-network classifiers from the literature to justify the use of neural networks for this task. We also briefly compare these results with state-of-the-art classifiers on the same dataset.

The MNIST dataset has become a standard benchmark for learning, classification and computer vision systems. Contributing to its widespread adoption are the understandable and intuitive nature of the task, the relatively small size and storage requirements and the accessibility and ease-of-use of the database itself \cite{7966217} . However, in 2017, EMINST \cite{7966217} was introduced as an extended MNIST dataset containing not only hand-written digits, but also all the letters in the English alphabet. 

We are going to encapsulate our experiments using a known variant of the EMINST dataset - Balanced EMNIST: out of 62 possible classes in EMNIST, we choose only 47. The motivation behind this is the fact that there are 15 letters for which it is confusing to discriminate between upper-case and lower-case. For the following letters, the labels were merged: C, I, J, K, L, M, O, P, S, U, V, W, X, Y, Z. This makes it more challenging to implement a well-rounded solution due to its higher complexity than the classic MNIST dataset, but also allows for richer comparisons among the methods implemented.

We explore this dataset by using multiple techniques. PCA for dimensionality reduction and reconstruction of the images; deviation from the mean vectors for each class; deviation from the original when doing reconstruction for sample reduction. Other recent studies like \cite{kim2020line} prove that sophisticated data mining methods (line-segment feature extraction) do work in reducing the dimensionality on this type of dataset that focus. Nevertheless, we show that even basic methods can work well if they are used properly, in a smart way. We provide visualizations of our motives and claims through the data mining process for a better understanding of the outcome of this paper.

There are 100k samples for training data, 15.8k for validation and 15.8k for testing dataset. These share the same image structure and parameters as the original MNIST task, allowing for direct compatibility with all existing classifiers and systems. Furthermore, the data is decently balanced; for train, each label has between 2076 and 2175 samples and for validation set, each label has between 297 and 380 samples. The input representation consists of a vectorized version of an 28x28 image, each pixel value will represent an entry in the input layer, so we will have 784 input units as the first layer. On disk, the training set amounts to approximatively 86 MB, however, we have reduced it to 76 MB for this task by saving the images in numpy array format. The same transformation was done for validation and test data beforehand.  For all our experiments, we chose stochastic training over 100 epochs for learning with Adam optimizer, mini-batch training with 100 samples per batch. Furthermore, for the weights initialization, we always choose from the same distribution.

The implementation for the neural networks is done in pure python with numpy for maximum computational efficiency, no frameworks were used here, everything was implemented from scratch. For the other models used for comparison reasons (Random Forest, Logistic Regression), we have used sklearn framework.

The motivation for this work is the abundance of research regarding convolution-based approaches in this field \cite{yoo2015deep, al2017review}, while the classic, standard MLP models have been rather forgotten. We want to show that these old approaches can still be a powerful tool in basic image classification at an exponential decrease in computation cost aiming to combat this shortage of MLP research in computer vision.

To sum it all up, there are 3 research questions that this paper is trying to answer to, through empirical work. First, how well can a pure fine-tuned MLP model classify hand-written digits and letters? Second, how much can we shrink the memory necessary to capture the dataset necessary for training using basic data mining techniques such that the accuracy performance doesn't drop significantly compared to training on the full unchanged dataset? Third, are other non-neural-network adaptive models capable of achieving similar or better performance than a fine-tuned MLP approach? 

We answer the first question employing a standard grid search over the hyper-parameter search space and analyzing the impact of two regularization techniques to boost the generalization performance of the model. For the second question, we define a non-significant drop in test accuracy as having a $<1\%$ difference. We respond to this by employing feature selection, dimensionality reduction and sample reduction classic approaches coming with a definite answer (a 67.2\% cutback). For the last question, we prove empirically that other models do not come close to the performance of a neural network in this computer vision task which might be the reason why neural networks, in general, have taken over the field of image classification \cite{al2017review}.

\section{Problem identification in baseline models}

We first tried a shallow architecture with vanilla gradient descent optimizer with a learning rate (lr) of 0.1 for mini-batch (100 samples) stochastic training, 100 neurons on the hidden layer and cross entropy Softmax error. We trained it over exactly 100 epochs, the results can be seen in figure 1:

As we can observe in figure 1, the validation error is higher than training error, moreover the validation error keeps increasing over time while the training error keeps decreasing. We can also see that the accuracy keeps going up for the training dataset, but stagnates in case of validation set. An ideal situation that we want to achieve is validation error to be low, slightly higher than the training error.

This is a well-known problem in training Neural Networks, caused by  $\textbf{overtraining}$. In our view, to further elaborate on the matter, this is because the Neural Network in this stage has the sole goal of minimizing train loss, it does not consider validation set at all, this is just a tool for us to assess the performance of the model during training, making sure it is not overfitting. However, this won't be an issue if the network finds the relevant information and patterns in the data, as if it does, the generalization would be good, hence the validation loss would continue to decrease. That would be ideal, but the network can also extensively keep trying to decrease the train loss function without any relevant plan, overfitting on the data failing to generalize well, therefore the validation error becomes a parabola shaped convex figure. 

\begin{figure}[H]
\begin{center}
\centerline{\includegraphics[width=0.6 \columnwidth]{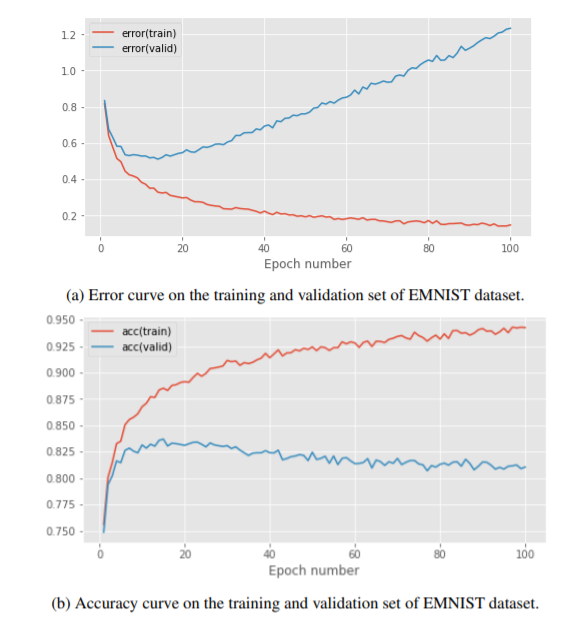}}
\caption{Accuracy and error curves on a baseline model on balanced EMNIST dataset.}
\end{center}
\end{figure} 

In general, though, there are multiple issues that can cause this: 

\begin{itemize}
  \item The distribution of the data is not the same for training and validation - the validation set might not be representative for the one that you train on. This usually happens when the data in not shuffled or it's unbalanced, as a solution we can try shuffling of the train and validation sets before training and also checking the distribution of the classes. This is not the reason for our algorithm as the data is shuffled and balanced.
  \item The model fails to decipher $\textbf{general patterns}$ in the data, maybe due to the fact that it's too simple. If the task is difficult, we can't expect it to be solved just by a model with very few parameters. The training error would keep decreasing until premature convergence and the validation error might be very far away. The accuracy on training set will keep increasing until stagnation, but the accuracy on the validation set will always be much lower and would possible remain steady in an interval after some point. $\textbf{This looks to be the case for our algorithm}$, it is pretty clear that it does not generalize well enough and it overfits on the training data, failing to find higher level patterns to succeed when queried with unseen data units. A solution would be trying a model with higher number of parameters or regularization.
  \item The model is too complex and it overfits on the training data causing the validation error to be far away behind or to stagnate while the training error keeps rising. After a certain point, it is possible for a model with very high number of parameters to learn everything by heart, hence not really showing signs of improvement when queried with unseen data units. A solution would be applying overfitting preventing methods like dropout; L1,L2 regularization.
  \item There are too many inputs for a simple model to handle and make sense of. This is pretty similar with the second bullet point, but it is obvious that our model, having only 100 neurons on the hidden layer might not be powerful enough to assess 784 input units (as the images are 28x28). Solutions for this include: partial feature selection done beforehand, trying a more complex model.
  \item For medium-sized datasets,  we can get unlucky even after checking the distribution and balancing the data, to train our model on easy samples to learn and validating mostly on hard cases. This happened to me in the past but it is generally very rare.
  \item The optimizer might not be suited for the problem or the final layer's activation is not suited for the problem. An optimizer based on GD and momentum might not be that great without a good learning rate, if it's too small the model might fall into a local minima and not generalize well for the validation set accuracy to get higher, it it's too high, the algorithm might diverge with continuous higher loss.
\end{itemize}

We can clearly see a point of $\textbf{minimum}$ for the validation loss in figure 1, that's where usually the $\textbf{generalization}$ is at its $\textbf{max}$, we could basically stop at around epoch ~$\textbf{10}$, because there is really no real $\textbf{improvement}$ afterwards.

In figure 1, we can also see that the error on the validation set increases on a very $\textbf{fast pace}$ but the accuracy on the validation set remains steady around 82\%. Why isn't it also decreasing at a fast pace? Basically, the model still succeeds in classifying the labels correctly, in most of the cases, but it is $\textbf{less and less sure}$ about it. If before, the model classified a real "one" as one with a confidence of 0.8, now it might classify it with a confidence of 0.55, that is still very high for a 47-labeled classification task, but it is a noticeable difference in loss.

It is pretty clear that the model does not generalize well and it may overfit on the training data as well. The most probable cause of the poor performance is the simple architecture of the current model. That's why we are going to try varying the number of hidden neurons and the number of hidden layers and observe what changes. We expect a higher number of model parameters (more hidden neurons) with regularization techniques to be more suited for our problem.
 
As a more solid baseline, we tried multiple shallow neural networks with 32, 64 and 128 neurons on the hidden layer. The optimizer used is Adam with a learning rate of 0.1 and as a loss function we use CrossEntropySoftmaxError.  This set-up is a classic one for a normal shallow neural network used for multi-class classification, the motivation behind choosing a learning rate of 0.1 has multiple points behind it: a higher learning rate with decay proved to be suited for the MNIST style datasets \cite{fcnch2018assessing}, \cite{anuradha2019assessment}; standard learning rate choices are 0.1, 0.01 and 0.001 - however, for only 100 epochs, a learning rate smaller than 0.01 would make the model too slow; the algorithm from figure 1 was trained using a learning rate of 0.1, so it is good to keep it for comparison reasons (if the algorithm does not diverge).
\begin{figure}[H]
\begin{center}
\centerline{\includegraphics[width=0.7 \columnwidth]{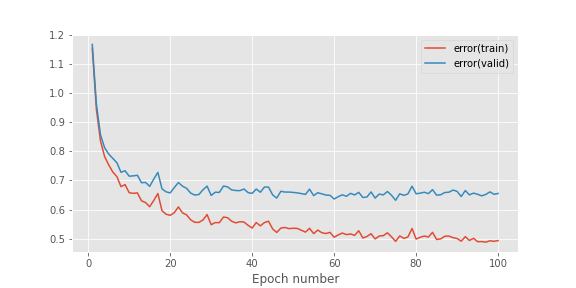}}
\caption{Shallow design, 32 hidden neurons}
\end{center}
\end{figure} 
\vspace{-2.5cm}
\begin{figure}[H]
\begin{center}
\centerline{\includegraphics[width=0.7 \columnwidth]{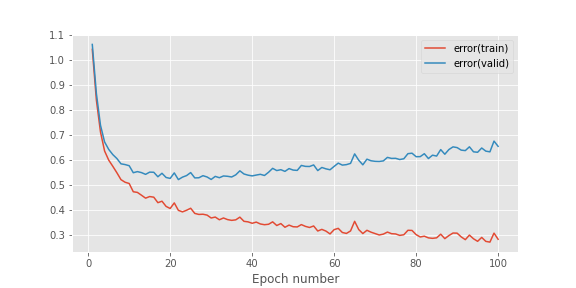}}
\caption{Shallow design, 64 hidden units}
\end{center}
\end{figure} 
\vspace{-2.5cm}
\begin{figure}[H]
\begin{center}
\centerline{\includegraphics[width=0.7 \columnwidth]{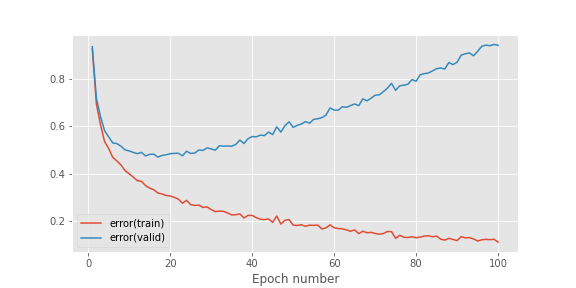}}
\caption{Shallow design, 128 hidden units}
\end{center}
\end{figure} 
\vspace{-2.5cm}

\begin{figure}[H]
\begin{center}
\centerline{\includegraphics[width=0.7 \columnwidth]{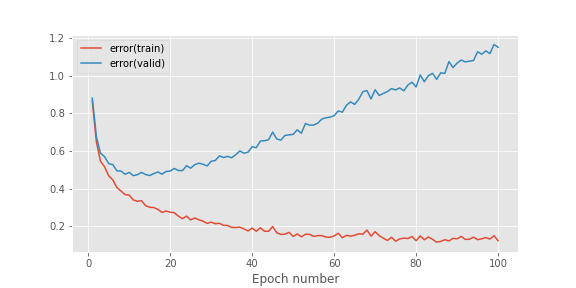}}
\caption{2 hidden layers, 128 neurons each, ReLu activations}
\end{center}
\end{figure} 

\begin{figure}[H]
\begin{center}
\centerline{\includegraphics[width=0.7 \columnwidth]{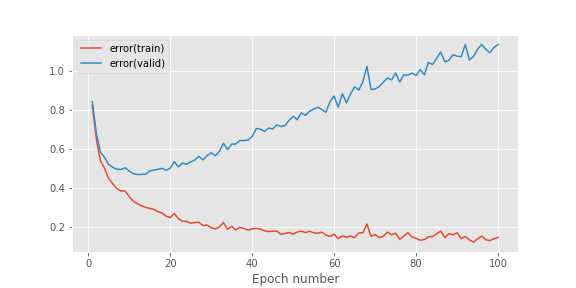}}
\caption{3 hidden layers, 128 neurons each, ReLu activations}
\end{center}
\vspace{-0.5cm}
\end{figure} 

The results (fig 2, 3, 4) show that the generalization gap gets worse and worse as we keep increasing the number of neurons. For $\textbf{32}$ neurons, the validation error stagnates but it does not increase, this might be as a consequence of the model having higher $\textbf{bias}$ and lower $\textbf{variance}$ when we have a lower number of hidden units. One other thing that is important to mention, although the $\textbf{32}$-units variant of the model does not suffer from overtraining that much compared to the other ones, the final and best validation accuracy stays at around $\textbf{0.8}$. Putting that into perspective, the best validation accuracy is around $\textbf{0.825}$ for $\textbf{64}$-units version and $\textbf{0.85}$ for the $\textbf{128}$-unit version. So if we were to choose the best model among these, the $\textbf{128}$-unit version, at the $\textbf{minimum point/inflexion point}$ when the validation error begins to rise up, has the most potential. That's why the following 2 experiments have been conducted starting with the 128 hidden units shallow network increasing the number of layers to 2 and 3.

Again (fig 5, 6), increasing the complexity of the model doesn't do anything but making the generalization gap higher and higher, the accuracy does not break the previous best of $\textbf{0.85}$ either. We also trained the models using a learning rate of $lr=\textbf{0.001}$ and the results were consistent, the graphs look almost identical. The neural network with 3 hidden layers has a lot of potential, but we need to fix the current issues by considering regularization techniques like $\textbf{dropout}$ - averaging multiple networks or $\textbf{L1, L2 regularization}$ to $\textbf{lower the variance}$ of the model by penalizing the weights.

\section{Dropout and Weight Penalty}
Dropout \cite{srivastava2014dropout} is a way of preventing overfitting in training neural networks with high number of model parameters. The main idea behind it is to randomly deactivate (drop) units along with their connections from a neural network during training. This way, the training stage is like using multiple neural networks that we \textit{average} in the end by considering smaller weights for testing. Intuitively, this is like forcing the flow of the neural network to find other paths, considering subsets of features in the feedforward stage - to reach the output and backpropagation stage - for learning.

\begin{algorithm}[ht]
\begin{algorithmic}
   \STATE {\bfseries Input:} data $inputs$, var $p$, boolean $evaluation$
   \IF{$evaluation = True$} 
   \STATE Return $p \cdot inputs$
   \ENDIF
  \STATE Initialize random matrix $\textit{mask}$ with dimensions like inputs
  \STATE Return  $(mask \le p) \cdot inputs$.
\end{algorithmic}
  \caption{Dropout Layer Forward Propagation}
  \label{alg:example}
\end{algorithm}

\begin{algorithm}[ht]
\begin{algorithmic}
   \STATE {\bfseries Input:} data $outputs$, var $grads$
   \STATE Return  $(outputs \neq 0) \cdot grads $
\end{algorithmic}
  \caption{Dropout Layer Backpropagation}
  \label{alg:example}
\end{algorithm}

To further elaborate on dropout, when a new mini-batch comes in, we disable with a probability $\textbf{p}$ a portion of the neurons on each hidden layer and use this set-up for forward and back propagation. There are multiple ways to balance out the fact that we are deactivating units in our network, we can train it as we described before and at only the testing stage, we can scale hidden units activations by $\textbf{p}$. Analogue, we could also scale the activations during training by 1 over the probability that the neuronal is included and let the testing phase untouched. Another technique worth mentioning is decay-ing the dropout probability to 0, in the same way we use it for learning rate - when we start with a large value that exponentially decays over time.

This is very effective as a regularization technique, because, as pointed out in the official paper \cite{srivastava2014dropout}, this is like training an exponential number of networks, all with different model parameters which means each one overfits in different ways, so most probably overfitting won't be a problem for each one individually. Moreover, this technique makes the average usefulness for each neuron increase, as if the important ones for the problems happen to be disabled, the network is forced to tune the other ones instead, to decrease the loss function. One can find similarities between this regularization method and ensemble learning \cite{dietterich2002ensemble}, with dropout, we combine models with high variance into one with a little higher bias but the overall variance decreases, it is like bagging.

Limitations of dropout consist in poor performance on convolutional layers that are the building blocks for CNNs \cite{huang2017densely, wide2016res, he2016deep}. Dropout shouldn't be actually used between convolution layers, but it is still very useful between dense layers, so this does not make it completely irrelevant in complex CNNs. The reason that it does not work is the fact that the weight gradients of the convolution layers are averaged across the spatial dimensions, which tend to contain many highly correlated activations, which is a problem with a different mask at every spatial location.

Weight penalty is another form of regularization in which we try to discourage the complexity of the model. It is important to note that, a model overfits on the training data when it actually also $\textbf{learns the noise}$ present in that dataset. That's why training error gets better and better but validation error remains constant of gets worse. The goal with L1, L2 regularization, as with other regularization techniques, is to learn the general patterns in the data ignoring the noise.

The naming L1 and L2 comes from L1 and L2 norms in mathematics. Given a vector $x \in R^n$ in the Euclidean space, the $\textit{p-norm}$ $||x||_p$ is defines as: 
\begin{center}
$||x||_p = (\sum ||x_i||_p)^{\frac{1}{p}}$. 
\end{center}
It follows that L1 norm incorporates a sum of absolute values and L2 norm incorporates  a sum of squares. Let's consider a general regression problem in which we use MSE as the loss function (where $y_i$ is the target value and $\hat{y_i}$ is the predicted output): 
\begin{center}
$MSE = \dfrac{1}{N} \cdot \sum (y_i - \hat{y_i})^2$. 
\end{center}
The model learnt by network can have any form, but to illustrate the formulas in a simpler manner, let's consider that it's represented by a third-degree polynomial: 
\begin{center}
$\hat{y_i} = f(x_i) = \theta_0 + \theta_1 x_i + \theta_2 x_i^2 + \theta_3 x_i^3$. 
\end{center}
A problem: we have a classification dataset that can be discriminated well by a linear model but the third-degree polynomial model might be too complex to fit such points in the euclidean space forming curved decision boundaries that follow each exact class distribution. We do not want to overfit like that; instead, what L1 and L2 regularization does is penalizing some weights making them 0 or really close to 0, hence becoming negligible. A successful L1 or L2 regularization would penalize weights $\theta_2, \theta_3$ transforming the classifier (before applying a squashing function) into:
\begin{center}
$\hat{y_i} = f(x_i) = \theta_0 + \theta_1 x_i + \cancel{\theta_2 x_i^2} + \cancel{\theta_3 x_i^3}$ 
\end{center}
that will capture well the main essence of the data and it won't involve the validation loss get higher over time.

But how do we actually penalize the weights that way? And how do we know which weights to penalize? Well, that's the learning algorithm job to discover, we will just tell him that some weights need to be low by forcing the sum of absolute values or the sum of squares of the model parameters to be low. In the loss function definition, we add those sums that represent L1, L2 normalizations -corresponding to the L1, L2 norms in mathematics:
\begin{center}
$LossL1 = \dfrac{1}{N} \cdot \sum (y_i - \hat{y_i})^2 + \lambda \sum |\theta_i|  $

$LossL2 = \dfrac{1}{N} \cdot \sum (y_i - \hat{y_i})^2 + \lambda \sum (\theta_i)^2 $
\end{center}

The $\lambda$ parameter is the regularization term which dictates how severe the penalization of the weights would be \cite{srivastava2014dropout}. If it's 0 then we reach the normal loss function, but if it's very large, then the majority of the weights will be close to 0.

The L1 and L2 regularization serve the same purpose but there are significant differences between them. The first thing that comes to mind are the outliers, of course squaring something big will make it even bigger, that's why L2 regularization is not that robust to outliers, L1 is much preferred in this case. Second of all, L2 usually does not make the weights 0, it can get really close to that but, if we want to completely ignore some model parameters, L1 is much preferred; in this way, we can say that L1 regularization has also built-in feature selection as it will choose by itself what input features are relevant in solving the task (by assigning insignificant input features with a zero weight). This is a double-edged sword though, because this way L1 may not learn complex patterns, L2 has the advantage from this point of view. Other notable differences include the fact that L2 regularization has a non-sparse solution and L1 regularization has multiple sparse solutions.

Comparing dropout and L1, L2 regularization, it is hard to say for sure which one would be better in general, it depends on the problem. But it is no secret that the dropout does much so more than L1, L2 norms: the fact that we are trying an exponential number of neural networks to solve the problem adds much more robustness to the model, but both of them can make some model parameters negligible or completely negligible. The dropout and L1, L2 regularization can also be used in combination, there are papers that show success using that \cite{srivastava2014dropout}. 


\section{Balanced EMNIST Experiments}
We want to study the implications of using Dropout and/or L1, L2 regularization over the baseline models presented in section 2 to obtain a new baseline model that we will use for our \textbf{data mining} experiments. For this, we first tested the dropout mechanic over a hyper-parameter search space and see what's the overall tendency of the results. 

\subsection{Dropout Experiments}

We used $\textbf{grid search}$ for identifying the optimal values for the dropout keep rate, the learning rate and the type of activations. The base architecture is 128 neurons spread across 3 hidden layers and we have chosen to apply dropout to the hidden layers, one that drops outputs from the first hidden layer and a second one that drops outputs from the second hidden layer. 

The motivation for this comes from our understanding that we don't need to apply dropout on each layer to get the best performance, a certain amount of regularization should suffice (in our view). We also decided to keep all the information from the input layer as we do not want to lose important features from the images (a high keep rate of 0.8 can still work). Dropout keep probabilities of 0.25, 0.5 and 0.75 were tested. It is important to mention that we used dropout before the affine layers and ReLu because of how it was recommended in the official paper \cite{srivastava2014dropout}. We have also tried a learning rate of 0.1 first, then 0.01 and then 0.5. Figure 7 clearly shows an improvement when comes to generalization, the error curves look exactly like in an ideal case: the validation one comes directly above the train one and it follows it throughout without fluctuations or unexpectedly going up.

\begin{figure}[ht]
\begin{center}
\centerline{\includegraphics[width=0.75\columnwidth]{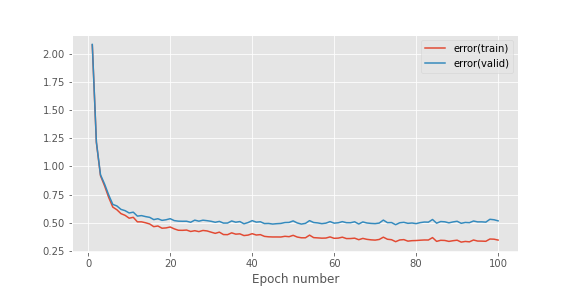}}
\caption{3 hidden layers, 128 neurons each, ReLu activations, Dropout p=0.5, lr=0.1}
\end{center}
\vspace{-0.5cm}
\end{figure} 

\begin{figure}[ht]
\begin{center}
\centerline{\includegraphics[width=0.75\columnwidth]{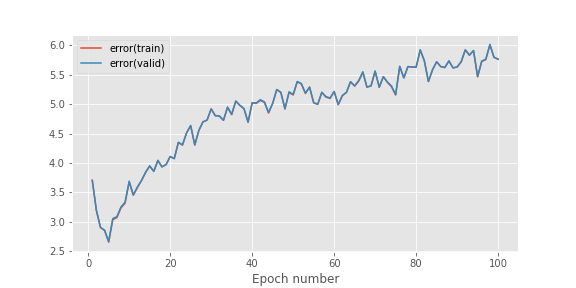}}
\caption{3 hidden layers, 128 neurons each, ReLu activations, Dropout p=0.25, lr=0.1 (Divergence)}
\end{center}
\vspace{-0.5cm}
\end{figure}

However, we did record a $\textbf{divergence}$ of the algorithm when it comes to the dropout keep rate of 0.25 (fig. 8), it is important to show failed cases as well. We thought this was because of the learning rate being too high but then we changed it to 0.01 and 0.001, the model still diverged, therefore it seems we are losing $\textbf{too much information}$ by only keeping a quarter of the units. This is also illustrated in \cite{srivastava2014dropout}, where this similar scenario happens and a value between 0.4 and 0.8 is recommended for dropout probability. Therefore, we can conclude that our architecture works well for relatively high dropout keep rates ($\ge 0.5$). 

This is expected - the authors of dropout also used a value of 0.5 in their experiments \cite{srivastava2014dropout}, though experimenting with this hyper-parameter is necessary. The results were very similar when using dropout keep rate of 0.5 and 0.75 when the learning rate varies below 0.1, so it seems a higher learning rate is ideal for faster training. In fact, a larger learning rate is also recommended in \cite{srivastava2014dropout}, appendix A, section 2, suggesting a value $\textbf{10}$ or $\textbf{100}$ times bigger than the one used on a standard neural network without dropout. This is $\textbf{intuitive}$ and it makes sense, as we do not have the same time to train every smaller neural network that results after dropout, we want to make the learning individually faster. A learning rate of 0.5 proved to be too large though as the validation error stagnates far above the train error, like in some of our baseline models. Differences between 0.5 and 0.75 as the dropout keep rate were seen in the final accuracies on the validation set, we got 85.2\% using the higher probability and 82\% with the probability of 0.5, so from this point of view, the higher keep rate would be ideal for this task.

\subsection{Weight Penalty Experiments}

When it comes to L1 regularization, this one also really shows that it improves the generalization strength of the model, however, we need to be very careful with picking the right regularization parameters. Our experiments show that the model diverges for high regularization parameters like 0.01, but does relatively well with regularization parameters in the range (1e-3, 1e-2). Going below 1e-3 with an order of magnitude  will not improve the baseline models suffering from the same problem. It's worth to note that a regularization parameter of 1e-3 worked best for us (fig 9), the model achieving almost 86\% validation accuracy which is a threshold that no baseline model achieved, but, compared to dropout, the curves look more $\textbf{noisy}$. Moreover, the validation error here dropped below 0.5, which no other model that we tested managed to achieve until now.

For L2 regularization, we decided to keep a learning rate of $\textbf{0.1}$ for all experiments and focus on the regularization parameter change as this learning rate proved to be good for standard dropout and L1 regularization. The results are very similar to the L1 norm, however, the regularization parameter needed changes as we found that an optimal value would be 1e-2 to really improve the generalization gap present in the baseline models. Training with this regularization parameter for L2 norm yielded the same exact best validation performance (fig 7). We also tried 1e-3, previous best value for L1, but the curves looked the same as the baseline models (with a slightly lower generalization gap). The same happened when we set the regularization parameter to 1e-4. However, with 0.01 as regularization parameter, we got an interesting result where the validation error and accuracy follow the train loss and accuracy really closely (fig 10), the difference being less than 0.25 for error and 0.1 for accuracy. All in all, it seems that L1 and L2 regularization parameters differ from one order of magnitude.

\begin{figure}[H]
\begin{center}
\centerline{\includegraphics[width=0.75\columnwidth]{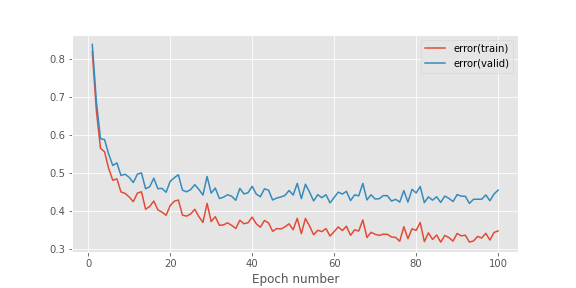}}
\caption{3 hidden layers, 128 neurons each, ReLu activations, L1 Regularization $\lambda=1e-4$}
\end{center}
\vspace{-0.5cm}
\end{figure} 

\begin{figure}[H]
\begin{center}
\centerline{\includegraphics[width=0.75\columnwidth]{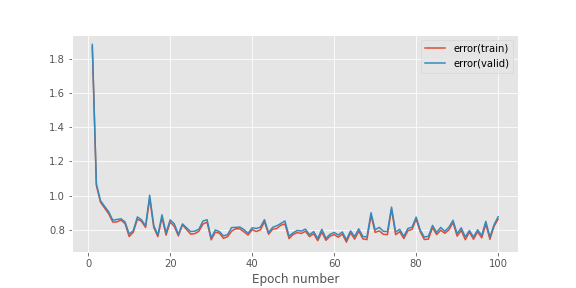}}
\caption{3 hidden layers, 128 neurons each, ReLu activations, L2 Regularization $\lambda=1e-2$}
\end{center}
\vspace{-0.7cm}
\end{figure}

\subsection{Hybrid Regularization}

We conducted extra experiments combining Dropout and Weight Penalty to see if we can get better results. $\textbf{In theory}$, we should select softer parameters for both these regularization techniques because, if applied simultaneously, there is always the threat of adding too much bias to the model - making too many assumptions about the data, killing its predictions. That's why, we expect out of the grid search hyper-parameter space, a high value of 0.75 for dropout keep probability and a small L1 regularization parameter of 1e-4 to work best. 

\begin{figure}[H]
\begin{center}
\centerline{\includegraphics[width=0.75\columnwidth]{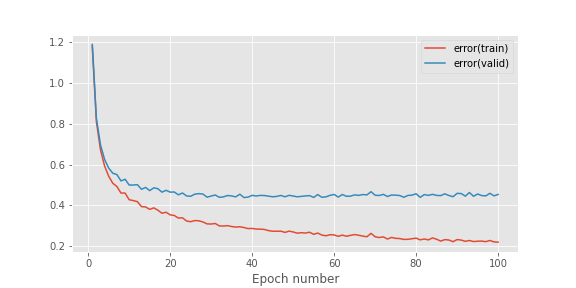}}
\caption{3 hidden layers, 128 neurons each, ReLu activations, Dropout p=0.75, L1 Regularization, coef= 1e-5}
\end{center}
\vspace{-0.7cm}
\end{figure} 

We tested only with L1 penalty as it will be equivalent in the performance with L2 with careful parameter setting. Again, the 0.1 learning rate was preserved. Of course, first, we tried with the previous best parameter settings, plus other combinations with 1e-2 and 1e-5 for L1 parameter, but in the end, as expected, the model with 0.75 keep probability for dropout and 1e-5 works best, achieving 85.7\% accuracy on the high end and 85.2\% on average after 50 epochs on the validation set. This is the best algorithm that we made with this kind of architecture  (figure 11). The best one using only L1 regularization is really close, however, this has less noisy-smoother performance as seen in the diagrams. The results of the majority of the experiments conducted ar present in table 1.

\begin{table}[H]
\begin{center}
        \centering
        \small
        \setlength\tabcolsep{2pt}
 \begin{tabular}{|c | c |c | c | c|} 
 \hline
 Hidden layers & Dropout & Weight Penalty & Train acc & Valid acc \\ [0.5ex] 
 \hline\hline
 1 (64 neurons) & No & No & 89.40\% & 81.60\% \\ 
 \hline
 1 (32 neurons)  & No & No & 83.20\% & 79.70\% \\
 \hline
 2 (128 neurons) & No & No &  95.20\% & 82.50\%\\
 \hline
 3 (128 neurons) & No & No & 94.40\% &  82.20\% \\
 \hline
 3 (128 neurons) & p=0.75 & No & 92.00\% & 85.20\% \\
 \hline
 3 (128 neurons) & p=0.50 & No & 84.20\% & 82.20\% \\
 \hline
 3 (128 neurons) & No & L1=1e-4 & 87.40\% & 84.60\% \\
 \hline
 3 (128 neurons) & No & L1=1e-5 & 94.40\% & 82.60\% \\
 \hline
 3 (128 neurons) & No & L2=1e-4 & 95.50\% & 83.80\% \\
 \hline
 3 (128 neurons) & No & L2=1e-3 & 87.60\% & 84.30\% \\
 \hline
 3 (128 neurons) & No & L2=1e-2 & 73.7\% & 72.9\% \\
 \hline
 3 (128 neurons) & No & L2=1e-5 & 94.10\% & 82.00\% \\
 \hline
 3 (128 neurons) & p=0.5 & L1=1e-4 & 71.40\% & 69.80\% \\
 \hline
 3 (128 neurons) & p=0.5 & L1=1e-5 & 87.50\% & 83.00\% \\
 \hline
 3 (128 neurons) & p=0.75 & L1=1e-5 & 91.40\% & 85.70\% \\ [1ex] 
 \hline
\end{tabular}
\end{center}
\caption{\label{tab:table-name} Results of the experiments so far.}
\end{table}
\vspace{-0.7cm}

\subsection{Summary}

All these experiments prove that regularization and dropout help to establish a better generalization performance over time, even though it seems that the model falls into $\textbf{local minima}$ - especially around the point where the validation loss hits 0.5.

Current state-of-the-art models for this dataset hit over 99\% accuracy \cite{anuradha2019assessment}, but they do use special architectures with deeper learning models and good feature selection. They can take dozens of hours to days to train on the same dataset making them awkward to use on less capable hardware. It is interesting to point out that some of these methods, like AlexNet or VGGNet, do get around 81-83\% accuracy when trained on half the dataset (around 50k samples) and GoogleNet gets around 88\% \cite{anuradha2019assessment}. Therefore, for comparable performance, we can say that more sophisticated methods need only half the dataset to achieve similar performance to vanilla end to end models, due to so much better feature selection.

On test dataset, our best model scored really close to the validation one, $\textbf{84.83\%}$ accuracy, which is stronger than the baseline results on this dataset present in \cite{7966217} but meets the overall expectation from the literature - topping at around 85\% validation accuracy and $\textbf{91.5\%}$ training accuracy. Moreover, the model took $\textbf{4.8s}$ per epoch which is really fast on just an i7 CPU. It is important to keep in mind that similar performance on validation set can be obtained with precise $\textbf{early stopping}$ without regularization, as our initial 3 layered neural net also scored high validation accuracy before stagnating, but in the long run (for higher epoch training), our tested methods of regularization will achieve higher accuracy on the test dataset.

\subsection{Applying Data Mining}

For such a non-convolutional architecture, the results are good enough and are in the expected range \cite{obradovic2018high, 7966217}. However, we do use a lot of memory space, just for training we have a 2D vector of shape 100k for the first dimension and 784 for the second dimension. On disk, with our actual numpy array representation, this amounts to approximatively \textbf{76 MB} of data just for training. With validation and test data, we reach roughly \textbf{100 MB}. The goal is to make this size smaller to fit on less capable hardware and to make training time faster without losing too much test accuracy performance. 

There are two intuitive ways this can be done. Reducting the feature space or reducting the number of samples we feed into the algorithm. Of course, convolutions do in a way realize feature extraction, but our main target is not to generate new features but to explore whether all of our current ones are really necessary to obtain top accuracy or not.

\subsubsection{Reducing the feature space}

That's why we have decided to do a PCA \cite{wold1987principal} on the training set and see how many principal components are needed to catch the most of the variation in the data. The cumulative explained variance ratio  \cite{wold1987principal} used to interpret that is shown in figure 12.

\begin{figure}[ht]
\begin{center}
\centerline{\includegraphics[width=\columnwidth]{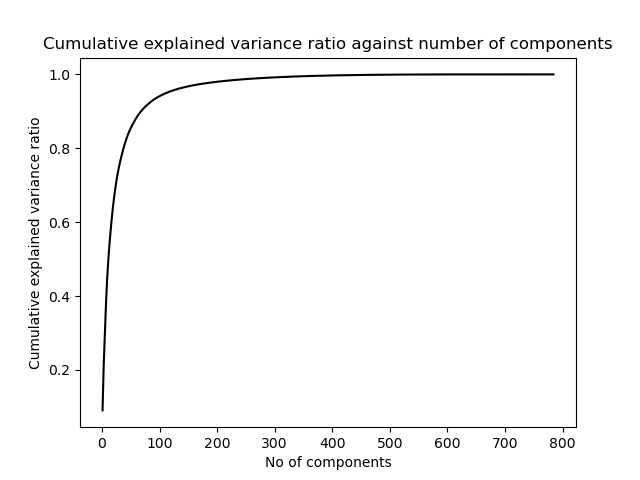}}
\caption{PCA cumulative explained variance ratio}
\end{center}
\vspace{-0.7cm}
\end{figure} 

This brings good news, the area below the graph is large, meaning, we need much fewer dimensions to capture the variance of the whole training dataset. With only \textbf{10\%} of the total number of components, we are able to catch \textbf{91.81\%}. of the variance. It is good practice to usually aim for \textbf{90\%} to \textbf{95\%} when doing dimensionality reduction with PCA, so if we project the dataset on the first 78 components we get a \textbf{90\%} reduction in the feature space. Plugin this into the best model we developed so far (table 1 configuration from last row), we actually get a slight increase in the test accuracy in the end to \textbf{85.08\%} (which is equivalent  to 40 more samples out of 15.8k classified correctly). Moreover, we get a \textbf{46.8\%} decrease in the time needed to train the same neural network. The validation accuracy gets slightly \textbf{higher} in the end as well, with \textbf{only 10\% of the original features}, as we can see in table 2. This reduction in dimensionality decreases the memory necessary on disk by \textbf{64\%} as well, which is a quite a step down in terms of size, if we choose to only work with this feature space. We need to keep in mind, though, that the PCA transformation matrix also needs to be saved in memory - it is being used every time we project a new data unit on the new feature space.

\begin{table}[H]
\begin{center}
        \centering
        \small
        \setlength\tabcolsep{2pt}
 \begin{tabular}{|c | c |c | c | c|} 
 \hline
 Hidden layers & Dropout & Weight Penalty & Valid acc & Duration \\ [0.5ex] 
 \hline\hline
 1 (32 neurons)  & No & No & 79.70\% & 120s\\
 \hline
 3 (128 neurons) & p=0.75 & No & 85.20\% & 340s \\
 \hline
 3 (128 neurons) & p=0.75 & L1=1e-5  & 85.70\% & 470s \\ 
 \hline
  \multicolumn{3}{|c|}{PCA and 3 hidden layers, p=0.75, L1=1e-5}  & 86.30\% & 250s \\
  \hline\hline
 \multicolumn{3}{|c|}{Random forest - 20 trees}  & 77.90\% & 25.54s \\ 
 \hline
 \multicolumn{3}{|c|}{Random forest - 50 trees} &  80.78\% & 68.89s \\ 
 \hline
 \multicolumn{3}{|c|}{Random forest - 100 trees}  & 81.62\% & 136.28s \\  
 \hline
 \multicolumn{3}{|c|}{Decision Tree}    & 50.5\% & 27.38s \\
 \hline
   \multicolumn{3}{|c|}{OneVsRest Logistic Regression}  & 67.63\% & 36.49s \\
 \hline
  \multicolumn{3}{|c|}{PCA and Random Forest 100 trees}   & 77.27\% & 173.96s \\
 \hline
  \multicolumn{3}{|c|}{PCA and Random Forest 50 trees} & 75.91\% & 91.17s \\
 \hline
  \multicolumn{3}{|c|}{PCA and Random Forest 20 trees} & 71.11\% & 33.24s \\ [1ex]
 \hline
\end{tabular}
\end{center}
\caption{\label{tab:table-name} Comparing multiple classifiers.}
\end{table}

In table 2 we can also observe the performance of other classifiers on this task. Of course, we can't really justify that a Neural Network is perfect for this task if we don't try other types of classifiers. We mainly tried Random Forest as it is usually a very powerful classifier that can rival neural networks in medium sized tasks. Support vector machines and KNN were also tried but the computational time necessary for these ones was incredibly high that it's not really feasible to use them here. \textbf{The accuracy obtained by the neural network is significantly higher than of the other classifiers, which justifies their use.} In fact, the performance of the Random Forest classifier with variable number of estimators is more comparable to what a shallow vanilla neural network with 32 hidden neurons would have. Granted, it is important to mention that the time spent on training is significantly less for the non-neural-network models on our machine, but this does not really motivate the big loss in accuracy.
%
\subsubsection{Reducing the sample size}

%
\begin{figure}[H]
\vspace{-0.7cm}
\begin{center}
\centerline{\includegraphics[width=0.7\columnwidth]{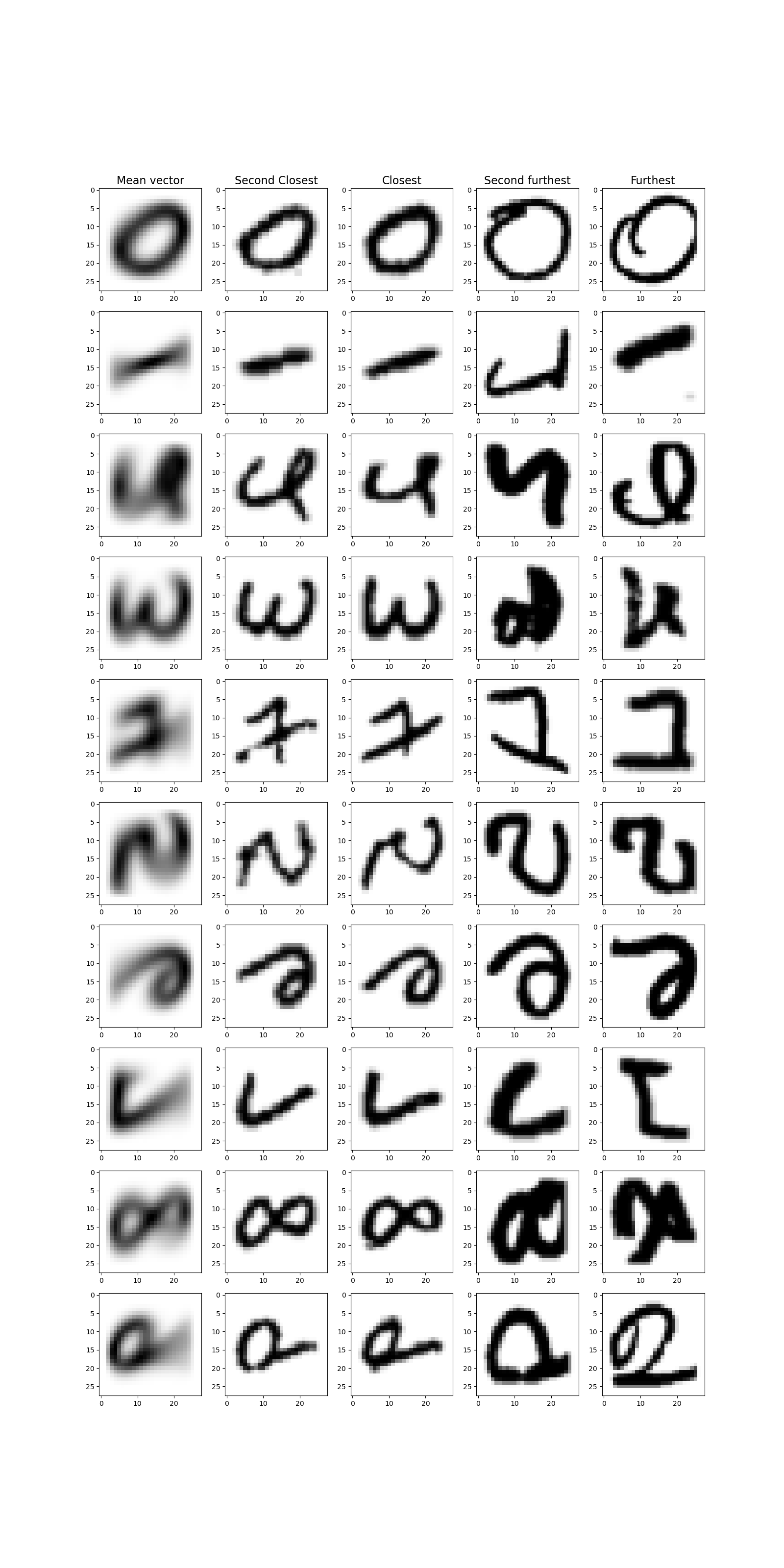}}
\vspace{-1.5cm}
\caption{Deviation from the mean for the first 10 classes}
\end{center}
\end{figure}

This part is trickier than the feature selection part, as large amounts of data are the most important part of a Neural Network success, so we should definitely expect a loss in the test accuracy if we blindly get rid of training samples. However, for this specific dataset, as it comprises handwritten digits and letters, we can expect weird shapes for some specific classes, due to the fact that people write in different ways. These can be detrimental to a neural network because outliers can affect the gradient flow significantly. To observe this phenomenon, we plotted the mean vector for the first 10 classes and the 2 closest and furthest away samples from this mean. We used Euclidean distance to quantify the distance between two 2D arrays.

The mean vectors (figure 13) look shady which is to be expected as we average the pixel values over multiple images. It is a clear difference, though, between the furthest away samples from each class and the mean vectors from each class. In same cases (digit 3 or 7), the shapes look so awkward even for a human to classify, therefore it should not be a bad idea to question the relevance of these samples for training the models.

\begin{figure}[H]
\begin{center}
\centerline{\includegraphics[width=\columnwidth]{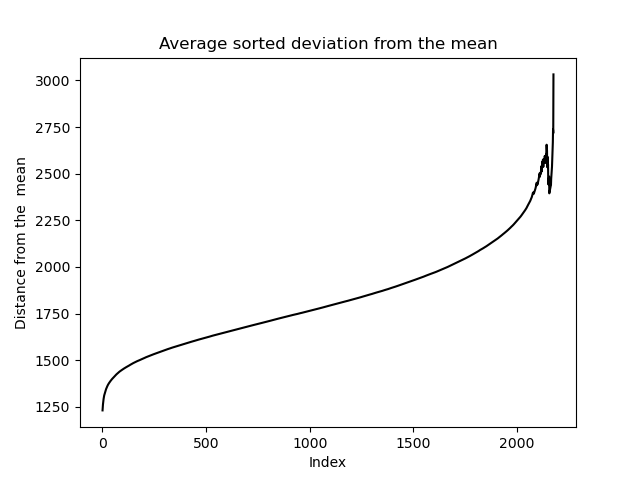}}
\caption{Distance from the mean}
\end{center}
\vspace{-0.7cm}
\end{figure}

An idea of removing training samples according to the deviation from the mean comes naturally, however, we need to establish a certain threshold for each class and anything that is above that gets excluded. To make a decision on that, we plotted the sorted average distance from the mean over all the classes (fig 14). We can see that the distance grows exponentially after the about 2000 sorted sample, so an idea of keeping only the closest 2000 samples from each class arises. This would mean a reduction of \textbf{6\%} in the training size, as we would have now only $47 \cdot 2000 = 94k$ samples, instead of 100k. Moreover, in this way, we can also perfectly balance the number of samples for each class during training.

Doing that for our experiment (and retraining a PCA model on the new training dataset), we get \textbf{84.29\%} test set accuracy, which is of course slightly lower than our previous best results but not that significantly. The validation accuracy we got also stayed in the same range: \textbf{85.2\%}. All in all, getting rid of 6 thousand images, in this way, did not cost the performance of the model that much and the training time is comparable to the previous best models. We have also tested the best configuration on different sizes of the training set reduced using the distance from the mean method. The results are present in table 3.

\begin{table}[H]
\begin{center}
        \centering
        \small
        \setlength\tabcolsep{2pt}
 \begin{tabular}{|c | c |c | c |} 
 \hline
 No  samples & No features  & Valid acc & Test acc \\ [0.5ex] 
 \hline\hline
 100k & 78 & 86.3\% & 85.08\% \\ 
 \hline
 94k  & 78 &  85.2\% &  84.29\% \\
 \hline
 89.3k & 78 & 84.7\% & 83.90\% \\ 
 \hline
 82k  & 78 &  83.1\% &  82.22\% \\
 \hline
 70k  & 78 &  79.9\% &  79.25\% \\
 \hline
 50k  & 78 &  74.2\% &  73.20\% \\ [1ex] 
 \hline
\end{tabular}
\end{center}
\caption{\label{tab:table-name} Results for the best model varying number of training samples.}
\end{table}

As we can see, if we use half the data through the minimum deviation from the mean method, we do get a \textbf{10\%} decrease in accuracy, which is a lot. However, at 94k and 82k the test accuracy is still over \textbf{82\%} which is impressive. We can say that for a \textbf{6\%} decrease in the number of samples, we get comparable performance with the case in which we use all the training samples.

This is not the only way to reduce the number of samples successfully. One can argue that we have used an encryption of the images with PCA to reduce the number of features; when doing the reconstruction, the newly formed image might or might not resemble that well the original image. Therefore, we have decided to compute the root mean square error, for each class, between the reconstructed samples and the original samples. This way, we can intelligently remove samples that are not a good reflection of the original data, because it can hurt the learning algorithm.  

We have plotted the sorted average RMSE over all classes (fig 15) to get an idea of the deviation from the original dataset.

\begin{figure}[H]
\begin{center}
\centerline{\includegraphics[width=\columnwidth]{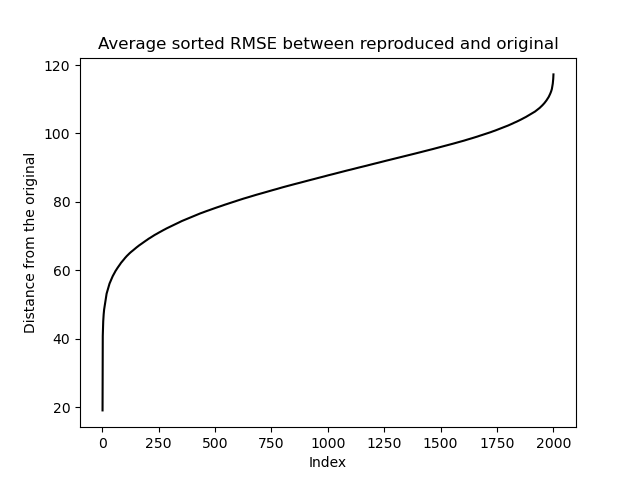}}
\caption{Deviation from the original - after removing 6k samples with the first method}
\end{center}
\vspace{-0.7cm}
\end{figure}

We can see that after around 1750 sorted samples, the deviation becomes more than 100 units and it grows exponentially after around 1900 samples. We tried multiple cut-off points for this as well and the results are present in table 4. 

\begin{table}[H]
\begin{center}
        \centering
        \small
        \setlength\tabcolsep{2pt}
 \begin{tabular}{|c | c |c | c |} 
 \hline
 No  samples & No features  & Valid acc & Test acc \\ [0.5ex] 
 \hline\hline
 89.3k & 78 & 84.6\% & 84.02\% \\ 
 \hline
 82k  & 78 &  83.2\% &  82.50\% \\
 \hline
 70k  & 78 &  80.5\% &  79.39\% \\
 \hline
 50k  & 78 &  72.5\% &  71.94\% \\ [1ex] 
 \hline
\end{tabular}
\end{center}
\caption{\label{tab:table-name} Results for the best model varying number of training samples.}
\end{table}

The results seem similar to when we were using only de deviation from the mean, they are a little bit better apart from the case when we reduced the training size to \textbf{50\%}. But with 10k samples less than the original size, we can get still over \textbf{84\%} test accuracy, which is impressive.

\subsection{Floating-point operations short analysis}

In this subsection, we want to examine the number of floating-point multiplication operations that are necessary for a forward method in our best multi-layer perceptron models. The main architecture that we use has 3 hidden layers with 128 neurons each; an input layer of 78 nodes and a Softmax Crossentropy output layer of 47 nodes. We can obtain the number of floating-point multiplication operations by calculating the following expression: $78 \cdot 128  + 128 \cdot 128 + 128 \cdot 128 + 128 \cdot 128 + 128 \cdot 47 =65,152$. This number is very low compared to what a CNN would have as number of operations (because it has to keep the 2D image - we can't do PCA in this case and because of the convolution computational cost). Nowadays, CNNs number of floating-point operations easily reach into the millions, but it strongly depends on the kernel on each layer and the number of kernels in each layer. To just give a quick intuition to the reader, let's image that googlenet \cite{szegedy2015going} had a sample from the Balanced EMNIST dataset as input: $28 \cdot 28$ and just one channel, everything else remaining the same (size of kernel $7 \cdot 7$ stride 2 padding 3, number of kernels for layer 1 is 64), then the number of multiplication operations needed for a forward pass through the first layer would be: $(\dfrac{28-7+3+3}{2}+1) \cdot (\dfrac {28-7+3+3}{2}+1) \cdot 7 \cdot 7 \cdot 64 = 614,656$. Note that here we used the well known formula $\dfrac{I-F+2P}{S}+1$  that defines the output shape after a convolution with kernel size $(F, F)$, stride $S$ and padding $P$. This is already one order of magnitude higher than in our case, for just one layer, which goes to show how expensive ConvNets really are compared to multi-layer perceptron models - vanilla neural networks.

\section{Related work}
Experiments with different variants of MLP in \cite{obradovic2018high, 7966217} set a literature expectation for this task and dataset asserting the performance on different variations of MLP.

In  \cite{7966217}, they used a 3-layered neural network as well, however, they make use of the Online Pseudo-Inverse Update Method (OPIUM) \cite{van2015online} for the weights update. No hyper-parameter-tuning was conducted. The  performance on the Balanced EMNIST dataset (testing set) obtained is 75.68\% with statistical error 0.05\%. This is clearly lower than our model but their main target was to provide an initial benchmark for further research.

In \cite{obradovic2018high}, they apply Hardware-aware training algorithms for this task. However, the architecture of the multi-layer perceptron model is a basic one - 3-layered neural network which seems similar to ours (before any processing step was applied). Not many details about the hyper-parameters are provided in the paper. Their test accuracy peaks below 85\%, which is what we can expect usually for this task, as we saw in our experiments as well.

Other studies have applied MLP models to EMNIST variations: \textit{Contrasting Convolutional Neural Network (CNN) with} \textit{Multi-Layer Perceptron (MLP)} \textit{for Big Data Analysis} \cite{botalb2018contrasting}. This study also compares the performance between a convolutional architecture and MLP. The dataset used is the  balanced EMNIST for characters only (26 classes instead of 47). We tried reconstructing the experiments from this particular paper to test how our own models would fair in a potential comparison. However, we haven't managed to provide a meaningful analysis because of the poor documentation of the cited study. On the EMNIST original paper, it is defined that the dataset for this 26-class balanced EMNIST for characters only has 145.6k samples, in total. This is contradictory to the cited paper's experiments where they use only about 29k samples, in total. It seems the authors have sampled a portion of the whole dataset without clear instructions for replication. Their best MLP model got 89.47\% after 202 epochs during training on all the dataset. Judging by the fact that the model has only a 5\% increase in performance with almost half the number of classes to be predicted will lean in our favour, however, we should keep in mind that a much smaller amount of samples have been used for training in their case, so we can't really pronounce  a definite statement on this matter. We should expect, though, that our algorithm will provide better accuracy performance as it has performed hyper-parameter tuning and data selection, which this study did not (for the MLP part). Interesting comparisons between the CNN and MLP can be observed in this study. The most impressive results here concern an analysis of the test accuracy after 15 epochs: while a CNN achieves over 90\%, the MLP model gets only 31.43\%. Out of curiosity, we have also interrogated the accuracy after 15 epochs of our best model (that obtained over 85\% accuracy on test set) and we get 83.9\%, which is decently close to the final one after 100 epochs. From a comparison point of view, this is not really that relevant because, again, we have more output classes, but we also have more samples to train on in contrast to the cited study.

Line-segment Feature Analysis Algorithm \cite{kim2020line} is a recent study that also focuses on hand-written digit and letter recognition and that also discusses PCA for the dimensionality reduction. However, its main contribution is a convolution-like feature extraction method that outperforms PCA in terms of final accuracy performance of a two machine learning models: KNN and SVM. The new algorithm extracts contours, lines, faces from the data and then identifies the types of line segments and sums them up \cite{kim2020line}. To extract features from line-segment information, LFA uses $3 x 3$ and $5 x 5$ filters. This is indeed an interesting study - we strongly consider implementing LFA to substitute the PCA part for a future work version of this paper. 

When dropout was introduced in 2014 as a way of tackling overfitting in deep neural networks \cite{srivastava2014dropout}, they also tested the effect on dropout on MNIST dataset.  This paper also evaluates dropout on a plethora of applications, ranging from vision and speech recognition to computational biology, as a method of improving model generalization alongside drastically reducing overfitting. It provides great practical recommendations for conducting our own experiments that we used in this paper. Although not that relevant to directly compare results, it shows strong empirical results in favor of dropout, especially  for MNIST dataset. However it does lack a theoretical underpinning that tackles the convergence of such a technique where its hyper-parameter vary. Dropout is still, to this day, relevant in fitting complex neural networks, but unfortunately, state-of-the-art models in vision recently adopted a residual convolutional architecture with other forms of regularization like BatchNorm \cite{ioffe2015batch} with rescaling and shifting that are simply better than dropout; although, granted, these appeared after the introduction of this technique.

\section{Discussion and Conclusions}

Vanilla dense neural networks architectures suffer from overfitting on the balanced EMNIST dataset. To mitigate the problem, we tested a lot of regularized models using dropout and L1, L2 norms that clearly show a massive improvement in generalization strength of the models over time, if fine hyper-parameter tuning is done. However, the algorithm still falls in local minima as our experiments' figures showed. The literature's solution for this is to use CNN models that use convolutions to extract spatial features from the data merged with dense regularized layers to add non-linearity to the tensors floating within in order to boost the accuracy on this 47-labeled classification task. That is because the model sometimes might struggle to extract relevant information from an image having only a flattened vector of 784 units (or 78) as input; better feature extraction can be done this way and surely will improve the results. This has been done already on MNIST dataset (\cite{simard2003best}, \cite{fcnch2018assessing}) and on EMNIST, with capsules, (\cite{anuradha2019assessment}) achieve more than 99\% accuracy. 

Our fine-tuned baseline model that uses no data mining beforehand gets a test accuracy of $\textbf{84.83\%}$, which is solid for this dataset considering the given architecture. It was obtained using dropout of 0.75 keep probability on the first 2 hidden layers, plus L1 penalty with a coefficient of 1e-5 on each layer, in a 3 hidden layered, 128 units each neural network design. This model takes up \textbf{470s} to train and the data size of disk necessary for training takes up \textbf{76 MB}. With a simple PCA, we deducted that we do not need all 784 dimensions for a good classification, changing the number of features from 784 to 78 actually boosted the test accuracy of the same model to  $\textbf{85.08\%}$. This newer model needs \textbf{250s} to train (\textbf{46.8\%} decrease) and only \textbf{28 MB} (\textbf{63\%} decrease) on disk for the training dataset. Finally, using the deviation from mean and the distance from the original samples through PCA inverse transformation to intelligently remove samples from the training set, we get comparable performance ($\textbf{84.02\%}$) with a further \textbf{10.7\%} decrease in terms of training data. This last model needs a little less than \textbf{250s} to train, but needs only \textbf{25 MB} of memory for the training dataset, which represents \textbf{32.8\%} of the total original training data size, roughly a third. This validates the importance of data minding in machine learning tasks.

\hfill \break

The following abbreviations are used in this manuscript:\\

\noindent 
\begin{tabular}{@{}ll}
MLP & Multi-layer Perceptron\\
EMNIST & Extended  Modified National Institute of Standards and Technology\\
SVM & Support Vector Machine \\
KNN & K Nearest Neighbours \\
ConvNets & Convolutional Neural Networks \\
CNN &  Convolutional Neural Networks \\
PCA & Principal Component Analysis \\
k & 1000 multiplier

\end{tabular}

\section{Declarations}

\subsection{Funding}

This research project did not receive any external funding.
\subsection{Availability of data and material}

The dataset is publicly available here: \href{https://www.nist.gov/itl/products-and-services/emnist-dataset}{link} (\textit{EMNIST Balanced} dataset). 
\subsection{Code availability}

The computer code is available at: \href{https://drive.google.com/drive/folders/1V-ExHR-8qKKa77DmSHo3uJNbf8_pFzto?usp=sharing}{link}. Step-by-step instructions to replicate the environment and run it properly are also present there. The experiments can already be visualized on the main jupyter notebook present in the mlp package.

\bibliographystyle{unsrt}

\small
\bibliography{bibliography}       

\end{document}